\documentclass[conference]{IEEEtran}
\IEEEoverridecommandlockouts

\usepackage{cite}
\usepackage{amsmath,amssymb,amsfonts}
\usepackage{graphicx}
\usepackage{textcomp}
\usepackage{xcolor}
\usepackage{booktabs}
\usepackage{url}
\usepackage{hyperref}
\hypersetup{colorlinks=true, linkcolor=blue, citecolor=blue, urlcolor=blue}

\def\BibTeX{{\rm B\kern-.05em{\sc i\kern-.025em b}\kern-.08em
    T\kern-.1667em\lower.7ex\hbox{E}\kern-.125emX}}

\begin{document}

\title{FusionCore: A 23-State Unscented Kalman Filter for\\
IMU, Wheel Encoder, GPS, and Visual SLAM Fusion in ROS~2}

\author{\IEEEauthorblockN{Manan Kharwar}
\IEEEauthorblockA{Independent Researcher, Hamilton, ON, Canada \\
manan.kharwar@outlook.com}
}

\maketitle

\begin{abstract}
We present FusionCore, an open-source ROS~2 sensor fusion package that fuses IMU, wheel encoder odometry, GPS, and Visual SLAM (VSLAM) pose into a single 100~Hz odometry stream using a 23-state Unscented Kalman Filter (UKF). The 23rd state is an online estimate of the wheel encoder's systematic yaw rate bias, which is identified through GPS heading cross-covariance and subtracted during GPS blackouts to reduce heading drift in coast mode. FusionCore also estimates gyroscope and accelerometer biases as explicit filter states, handles GPS natively in ECEF without a separate coordinate projection node, applies per-sensor Mahalanobis chi-squared outlier gating calibrated to measurement degrees of freedom, and adapts sensor noise covariance automatically from the innovation sequence. VSLAM pose fusion enables GPS-denied operation with any visual odometry or SLAM system, including automatic recovery from map reinitialization. We evaluate against \texttt{robot\_localization} on twelve full-length sequences (55--92~min each) from the NCLT public dataset \cite{nclt}. FusionCore achieves lower Absolute Trajectory Error (ATE) on ten of twelve sequences, with improvements ranging from 1.2$\times$ to 22.2$\times$ on winning sequences. The \texttt{robot\_localization} UKF diverges numerically on all twelve sequences. FusionCore is available at \url{https://github.com/manankharwar/fusioncore} under the Apache~2.0 license.
\end{abstract}


%------------------------------------------------------------------------
\section{Introduction}

Accurate pose estimation is a prerequisite for autonomous robot navigation. Fusing IMU, wheel odometry, and GPS is a well-studied problem, yet practical deployment remains difficult. IMU gyroscope and accelerometer biases drift continuously and must be estimated online. GPS arrives at low rate with intermittent quality degradation. Wheel odometry provides high-rate relative motion but accumulates error on slippery surfaces. Critically, the wheel encoder's own yaw rate bias (a systematic offset in differential drive angular velocity estimation) is rarely treated as a filter state, leaving it unobservable and free to accumulate heading error during GPS blackouts.

\texttt{robot\_localization} \cite{robot_localization} is the dominant ROS package for this task, with widespread adoption across academic and commercial robots. It offers an EKF and UKF, handles multiple simultaneous sensor inputs, and has extensive documentation. However, several architectural choices limit its utility for modern deployments: gyro and accelerometer biases are not part of the filter state; GPS requires a separate \texttt{navsat\_transform} node projecting coordinates through UTM, which introduces zone boundary failures; outlier rejection uses manually tuned Mahalanobis scalar thresholds that do not account for measurement dimensionality; and noise covariance is fixed at configuration time.

FusionCore addresses each of these limitations in a single ROS~2 lifecycle node with no additional coordinators. The contributions of this work are:

\begin{enumerate}
  \item A 23-state UKF formulation with continuous online gyroscope bias, accelerometer bias, and wheel encoder yaw rate bias estimation.
  \item ECEF-native GPS fusion eliminating UTM projection and its associated zone boundary failures.
  \item Per-sensor Mahalanobis chi-squared outlier gating with thresholds calibrated per measurement path and degrees of freedom.
  \item Automatic noise covariance adaptation from a sliding innovation window using exponential moving average.
  \item Automatic zero-velocity updates (ZUPT) that suppress bias drift during stationary periods.
  \item IMU ring-buffer retrodiction for GPS and VSLAM delay compensation.
  \item Visual SLAM pose fusion for GPS-denied operation, accepting \texttt{nav\_msgs/Odometry} from any visual odometry or SLAM system, with automatic recovery from map reinitialization.
  \item GPS velocity fusion via a configurable topic accepting \texttt{nav\_msgs/Odometry} for cross-sensor slip detection.
  \item Radar Doppler ego-velocity fusion via a configurable body-frame velocity topic for GPS-aided platforms with 4D imaging radar.
  \item A quantitative evaluation on twelve full-length sequences of the NCLT public dataset against \texttt{robot\_localization} EKF and UKF.
\end{enumerate}

%------------------------------------------------------------------------
\section{Related Work}

Kalman filtering for inertial-GPS fusion has been studied for decades \cite{groves2013}. Loosely coupled integration treats GPS as position measurements, while tightly coupled integration fuses raw pseudoranges. FusionCore implements loosely coupled fusion using \texttt{sensor\_msgs/NavSatFix} or, optionally, \texttt{gps\_msgs/GPSFix} for RTK-capable receivers.

\texttt{robot\_localization} \cite{robot_localization} is the canonical ROS implementation. It supports EKF and UKF with arbitrary sensor combinations. Known limitations include UTM zone boundary crashes \cite{rl_utm_issue}, numerical UKF instability under aggressive maneuvers \cite{rl_ukf_issue}, and the two-node GPS architecture requiring \texttt{navsat\_transform}.

MSF \cite{msf} and ROVIO \cite{rovio} target visual-inertial odometry (VIO). imu\_filter\_madgwick \cite{imu_filter} provides complementary filter orientation estimation but not full state estimation with GPS. GTSAM \cite{gtsam} and Ceres \cite{ceres} offer factor-graph optimization but require batch or sliding-window formulations not well-suited to online 100~Hz output. FusionCore targets the practical ROS~2 use case: a single lifecycle node that a practitioner can drop into an existing Nav2 stack with a YAML configuration file.

%------------------------------------------------------------------------
\section{System Architecture}

FusionCore is structured as two packages. \texttt{fusioncore\_core} is a pure C++17 library with no ROS dependency implementing the UKF, measurement models, and sensor abstractions. \texttt{fusioncore\_ros} wraps it as a ROS~2 lifecycle node, managing subscriptions, TF broadcasting, and lifecycle state transitions. This separation allows standalone use of the filter library in non-ROS environments.

The node subscribes to \texttt{sensor\_msgs/Imu} (primary IMU), an optional second IMU via \texttt{imu2.topic}, \texttt{nav\_msgs/Odometry} (wheel encoders, VSLAM pose, GPS velocity, and radar Doppler in body frame), and GPS position as either \texttt{sensor\_msgs/NavSatFix} or \texttt{gps\_msgs/GPSFix} (selectable at runtime). The GPSFix interface unlocks RTK\_FLOAT fix-type reporting and receiver-native HDOP/VDOP fields unavailable in the NavSatFix message. All sensor inputs except the primary IMU are optional; FusionCore supports any combination. It publishes \texttt{nav\_msgs/Odometry} at 100~Hz on \texttt{/fusion/odom} and broadcasts the \texttt{odom~$\to$~base\_link} transform.

%------------------------------------------------------------------------
\section{Filter Formulation}

\subsection{State Vector}

The filter maintains a 23-dimensional state vector:
\begin{equation}
\mathbf{x} = \begin{bmatrix}
\mathbf{p} & \mathbf{q} & \mathbf{v} & \boldsymbol{\omega} & \mathbf{a} & \mathbf{b}_g & \mathbf{b}_a & b_{\text{ewz}}
\end{bmatrix}^\top
\end{equation}

where $\mathbf{p} \in \mathbb{R}^3$ is position in ENU, $\mathbf{q} = [q_w, q_x, q_y, q_z]$ is unit quaternion orientation, $\mathbf{v} \in \mathbb{R}^3$ and $\boldsymbol{\omega} \in \mathbb{R}^3$ are body-frame linear and angular velocity, $\mathbf{a} \in \mathbb{R}^3$ is body-frame linear acceleration, $\mathbf{b}_g \in \mathbb{R}^3$ is gyroscope bias, $\mathbf{b}_a \in \mathbb{R}^3$ is accelerometer bias, and $b_{\text{ewz}} \in \mathbb{R}$ is the wheel encoder angular velocity Z bias.

Gyroscope and accelerometer biases are first-class filter states with dedicated process noise parameters $q_{\text{gyro\_bias}}$ and $q_{\text{accel\_bias}}$. This allows the filter to converge on sensor-specific bias values during operation without a separate calibration phase.

The 23rd state, $b_{\text{ewz}}$, is the systematic yaw rate bias of the wheel encoder. In a differential drive robot, this arises from wheel radius mismatch, mechanical asymmetry, and surface irregularities; it is a near-constant offset in the reported $\omega_z$. FusionCore identifies $b_{\text{ewz}}$ online through the cross-covariance between encoder $\omega_z$ error and the heading implied by GPS bearing: when GPS is available, the filter back-propagates observed heading error into the encoder bias state. During GPS blackouts (coast mode), the estimated $b_{\text{ewz}}$ is subtracted from encoder angular velocity before integration, reducing heading drift proportional to coast duration. This mirrors exactly how $\mathbf{b}_g$ handles gyroscope $\omega_z$ bias.

\begin{figure}[h]
  \centering
  \includegraphics[width=\columnwidth]{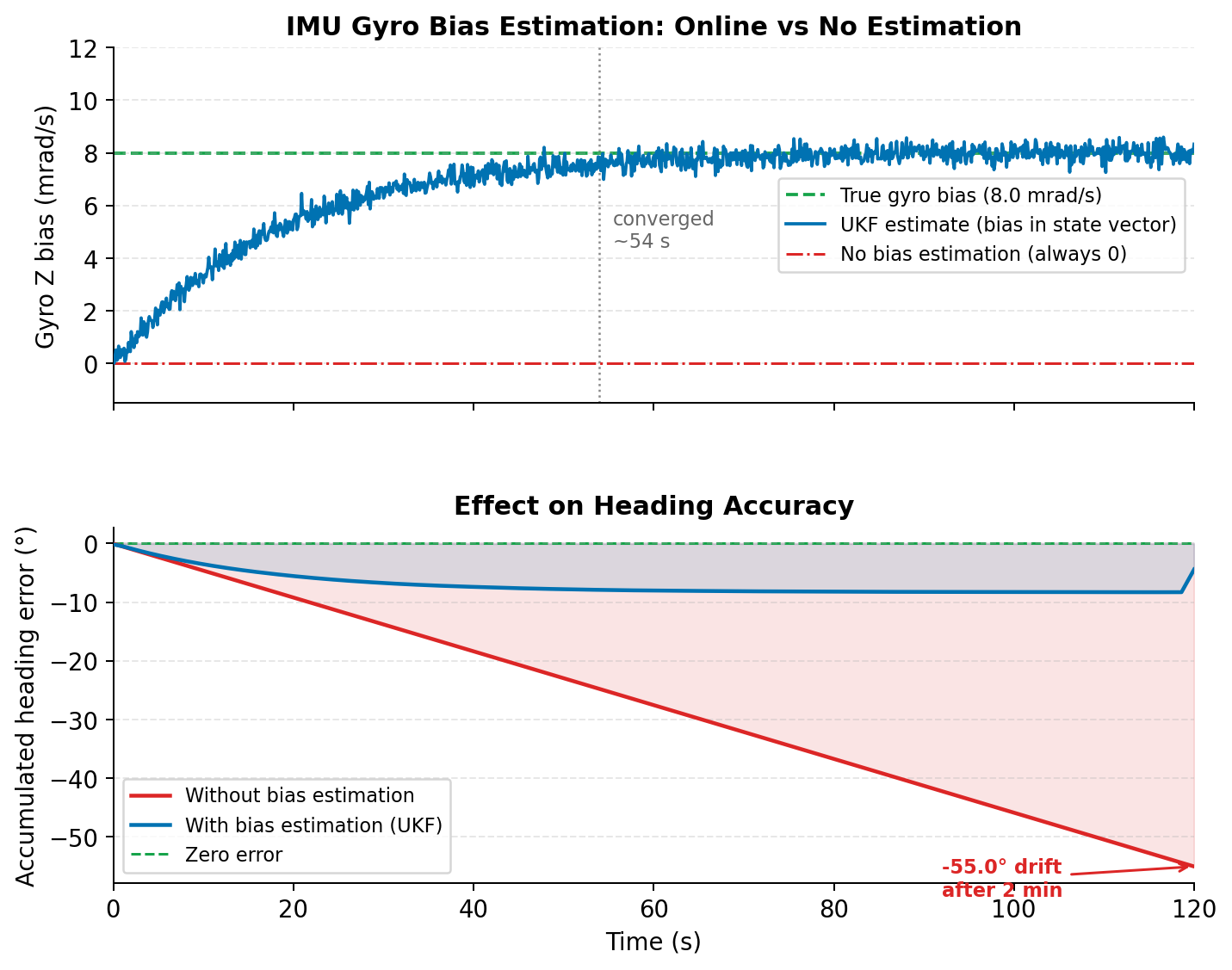}
  \caption{Online gyroscope bias estimation. An 8~mrad/s bias converges within 40~s. Without estimation, the accumulated heading error reaches $-55^\circ$ in 2~minutes; with UKF bias states, drift remains bounded.}
  \label{fig:bias}
\end{figure}

\subsection{Process Model}

The continuous-time process model is discretized at each IMU step ($\Delta t \approx 10$~ms):

\begin{align}
\mathbf{p}_{k+1} &= \mathbf{p}_k + \Delta t \cdot R(\mathbf{q}_k) \, \mathbf{v}_k \\
\mathbf{q}_{k+1} &= \mathbf{q}_k \otimes \exp\!\left(\tfrac{1}{2} \boldsymbol{\omega}_k \Delta t\right) \\
\mathbf{v}_{k+1} &= \mathbf{v}_k + \Delta t \cdot \mathbf{a}_k \\
\boldsymbol{\omega}_{k+1} &= \boldsymbol{\omega}_k, \quad
\mathbf{a}_{k+1} = \mathbf{a}_k \\
\mathbf{b}_{g,k+1} &= \mathbf{b}_{g,k} + \mathbf{w}_{bg} \\
\mathbf{b}_{a,k+1} &= \mathbf{b}_{a,k} + \mathbf{w}_{ba} \\
b_{\text{ewz},k+1} &= b_{\text{ewz},k} + w_{ewz}
\end{align}

Biases follow random walks driven by zero-mean Gaussian process noise: $\mathbf{w}_{bg} \sim \mathcal{N}(0, Q_{bg})$, $\mathbf{w}_{ba} \sim \mathcal{N}(0, Q_{ba})$, $w_{ewz} \sim \mathcal{N}(0, q_{ewz})$, with tunable parameters \texttt{ukf.q\_gyro\_bias}, \texttt{ukf.q\_accel\_bias}, and the encoder bias process noise respectively. This allows the filter to slowly re-estimate biases as they drift with temperature or operating conditions, rather than locking them at initialization.

The quaternion exponential for exact kinematics is:
\begin{equation}
\exp\!\left(\tfrac{1}{2}\boldsymbol{\omega}\Delta t\right) =
\begin{cases}
\left[\cos\theta,\; \frac{\sin\theta}{\|\boldsymbol{\omega}\|}\boldsymbol{\omega}\right] & \|\boldsymbol{\omega}\| > \epsilon \\
\left[1,\; \tfrac{1}{2}\boldsymbol{\omega}\Delta t\right] & \text{otherwise}
\end{cases}
\end{equation}
where $\theta = \tfrac{1}{2}\|\boldsymbol{\omega}\|\Delta t$.

\subsection{Sigma Point Generation}

The UKF uses $2n+1 = 47$ sigma points for the $n=23$ state. Three numerical issues specific to quaternion states require explicit handling:

\textbf{Quaternion sign consistency.} Two sigma points $\mathbf{q}$ and $-\mathbf{q}$ represent the same rotation but their weighted sum does not. We flip sigma points in the opposite hemisphere before averaging: if $\mathbf{q}_0 \cdot \mathbf{q}_i < 0$, negate $\mathbf{q}_i$.

\textbf{Lost positive-definiteness.} High-rate IMU updates cause the $KSK^\top$ subtraction to occasionally lose positive-definiteness in bias dimensions. We detect and repair via eigenvalue shift:
\begin{equation}
P \leftarrow P + (-\lambda_{\min} + \epsilon) I
\end{equation}
where $\lambda_{\min}$ is the minimum eigenvalue of $P$. Identity shift preserves eigenvectors; clamping individual eigenvalues does not.

\textbf{Angular velocity variance cap.} Without gyroscope measurement updates, $P(\omega_x, \omega_x)$ grows unboundedly through process noise. With Wm$_0 \approx -99$ (at $\alpha=0.1$, $n=23$), unbounded growth amplifies floating-point asymmetry into quaternion drift. We cap angular velocity variance at 1.0~rad$^2$/s$^2$, which is physically large but finite.

%------------------------------------------------------------------------
\section{Measurement Models}

\subsection{IMU}

Each IMU message produces two sequential updates.

\textbf{Raw IMU (6-DOF).} Gyroscope rates and accelerometer readings are fused jointly. The measurement function accounts for bias and gravity:
\begin{align}
h_\omega(\mathbf{x}) &= \boldsymbol{\omega} + \mathbf{b}_g \\
h_a(\mathbf{x}) &= \mathbf{a} + \mathbf{b}_a + R(\mathbf{q})^\top \mathbf{g}
\end{align}
where $\mathbf{g} = [0, 0, 9.80665]^\top$~m/s$^2$ in ENU.

\textbf{Orientation (2-DOF or 3-DOF).} For 9-axis IMUs with magnetometer, roll, pitch, and yaw are fused (3-DOF). For 6-axis IMUs, only roll and pitch are fused (2-DOF), leaving the yaw unobservable from IMU alone. This distinction prevents corrupting the GPS-observable yaw estimate with an inconsistent magnetometer measurement.

\subsection{Wheel Encoders}

Encoder odometry provides a 3-DOF velocity measurement. The encoder bias $b_{\text{ewz}}$ is subtracted from the reported $\omega_z$ before constructing the measurement:
\begin{equation}
h_{\text{enc}}(\mathbf{x}) = [v_x,\; v_y,\; \omega_z - b_{\text{ewz}}]^\top
\end{equation}
in body frame. Two additional pseudo-measurements enforce the non-holonomic ground constraint on every encoder update. The first, $h_{vz}(\mathbf{x}) = v_z = 0$, constrains body-frame vertical velocity. The second, $h_{az}(\mathbf{x}) = a_z = 0$, constrains body-frame vertical acceleration. The AZ constraint is necessary because a small mismatch between the IMU's local gravity reading and the WGS84 constant (9.80665~m/s$^2$) leaks into the $a_z$ state via large process noise; $a_z$ then integrates into $v_z$ through the motion model, so the VZ constraint alone cannot fully compensate. Both constraint noise values adapt online from the innovation sequence, automatically loosening when the robot traverses curbs or rough terrain and tightening back to configured floor values on flat ground.

\subsection{GPS}

GPS position is converted from WGS84 latitude/longitude/altitude to the filter's ENU frame using the PROJ library \cite{proj}. The reference origin is set at the first GPS fix received after activation. ECEF-to-ENU conversion avoids UTM zone boundaries entirely.

Noise covariance is constructed from HDOP and VDOP when a full 3$\times$3 covariance matrix is unavailable:
\begin{equation}
R_{\text{GPS}} = \text{diag}\!\left(\sigma_{\text{xy}}^2 \cdot \text{HDOP}^2,\; \sigma_{\text{xy}}^2 \cdot \text{HDOP}^2,\; \sigma_z^2 \cdot \text{VDOP}^2\right)
\end{equation}
where $\sigma_{\text{xy}}$ and $\sigma_z$ are user-configured baseline noise at HDOP/VDOP=1. Fix quality is gated by configurable minimum fix type (GPS, DGPS, RTK\_FLOAT, RTK\_FIXED), maximum HDOP, and minimum satellite count.

When using \texttt{gps\_msgs/GPSFix} (set via \texttt{gnss.use\_gps\_fix: true}), the filter reads \texttt{satellites\_used} directly, uses receiver-native \texttt{hdop}/\texttt{vdop}, and accepts \texttt{err\_horz}/\texttt{err\_vert} (95\% CI bounds) as an alternative covariance source. This unlocks RTK\_FLOAT status (fix type 3), which is unreachable via \texttt{sensor\_msgs/NavSatFix} whose status field maps only SBAS/WAAS, GBAS, and RTK\_FIXED.

GPS antenna lever arm correction is applied when heading is independently validated:
\begin{equation}
\mathbf{p}_{\text{antenna}} = \mathbf{p}_{\text{base}} + R(\mathbf{q})\, \ell
\end{equation}

\subsection{GPS Velocity}

Doppler-derived velocity from a GNSS receiver is fused as a 2-DOF ENU velocity measurement via \texttt{gnss.velocity\_topic} (a \texttt{nav\_msgs/Odometry} topic with \texttt{linear.x} = east, \texttt{linear.y} = north). This provides an independent velocity observation not dependent on wheel contact, enabling slip detection and cross-sensor consistency checking.

\subsection{Radar Doppler Velocity}

Body-frame ego-velocity from a 4D imaging radar is fused via \texttt{radar.velocity\_topic}. The measurement model is a 2-DOF body-frame velocity update (\texttt{linear.x} = forward, \texttt{linear.y} = lateral), structurally identical to the encoder model but independent of wheel contact. This is particularly effective for outdoor platforms at speed where wheel slip is difficult to detect from encoder data alone.

\subsection{Visual SLAM Pose}

VSLAM pose measurements are accepted as \texttt{nav\_msgs/Odometry} on a configurable topic. The 6-dimensional measurement function is:
\begin{equation}
h_{\text{vslam}}(\mathbf{x}) = [x,\; y,\; z,\; \phi,\; \theta,\; \psi]^\top
\end{equation}
where $\phi, \theta, \psi$ are roll, pitch, and yaw extracted from the filter's quaternion state. The yaw innovation is wrapped across the $\pm\pi$ boundary to avoid discontinuities. Per-message covariance from the SLAM system is used directly when available; diagonal elements are floored at $\sigma_{\text{pos}} = 1$~cm and $\sigma_{\text{orient}} = 0.001$~rad to prevent degenerate noise estimates.

VSLAM systems can reinitialize to a new map after tracking loss, producing a sudden large innovation that the chi-squared gate correctly rejects. FusionCore counts consecutive gate rejections on the VSLAM path; after a configurable number of consecutive rejections (\texttt{vslam.reinit\_n}, default 10), the map-to-odom transform offset is re-anchored to the filter's current position, recovering continuous operation without external intervention.

%------------------------------------------------------------------------
\section{Filter Design Choices}

\subsection{Outlier Rejection}

Every measurement undergoes a Mahalanobis chi-squared gate before modifying filter state:
\begin{equation}
d^2 = \boldsymbol{\nu}^\top S^{-1} \boldsymbol{\nu} \leq \tau_s
\end{equation}
where $\boldsymbol{\nu}$ is the innovation, $S$ is the innovation covariance, and $\tau_s$ is a configurable per-sensor threshold. Default thresholds and their chi-squared interpretations are: GPS position (3-DOF): 16.27 [$\chi^2(3,\, 0.999)$]; VSLAM pose (6-DOF): 22.46 [$\chi^2(6,\, 0.999)$]; heading (1-DOF): 10.83 [$\chi^2(1,\, 0.999)$]; encoder velocity (3-DOF): 11.34 [$\chi^2(3,\, 0.99)$]; IMU (all measurement paths): 15.09. The IMU threshold is applied uniformly across the 6-DOF raw update, 3-DOF orientation update, and 2-DOF orientation update paths; a single configurable parameter is used because the IMU is the highest-rate sensor and per-path tuning adds fragility. The value 15.09 corresponds approximately to $\chi^2(6,\, 0.98)$ and is intentionally more permissive on lower-DOF IMU paths, trading some specificity for filter robustness under aggressive maneuvers. All thresholds are configurable YAML parameters; the defaults were chosen to be permissive for sensors that produce rejectable measurements infrequently (GPS, VSLAM) and tighter for the high-rate IMU and encoder paths where individual rejections are inexpensive.

\begin{figure}[h]
  \centering
  \includegraphics[width=\columnwidth]{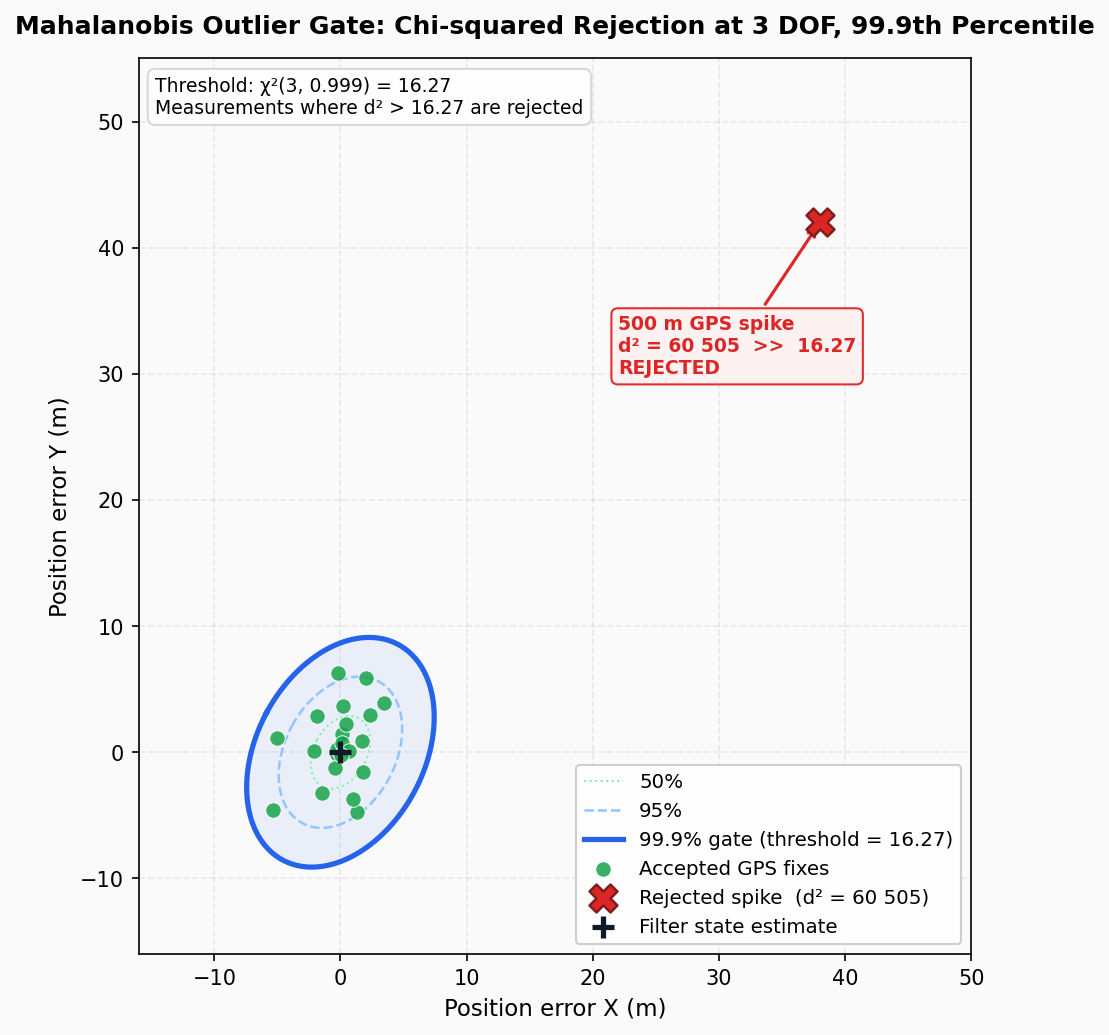}
  \caption{Chi-squared gate for GPS position (threshold 16.27, $\chi^2(3,\, 0.999)$). A 500~m GPS spike produces $d^2 = 60{,}505$, far beyond the threshold. The filter state is unchanged.}
  \label{fig:mahalanobis}
\end{figure}

\texttt{robot\_localization} exposes scalar Mahalanobis thresholds that users compare against the square root of $d^2$. This conflates measurement dimensionality: a threshold of 4.0 is appropriate for a 3-DOF measurement but overly permissive for a 1-DOF measurement. FusionCore pre-calibrates default thresholds per measurement path, and provides separate YAML parameters for each sensor so practitioners do not need to reason about chi-squared distributions to configure the filter.

\subsection{Adaptive Noise Covariance}

Sensor noise covariance $R$ is adapted online from the innovation sequence using a sliding window of 50 samples and exponential moving average:
\begin{equation}
R \leftarrow (1 - \alpha) R + \alpha \hat{C}
\end{equation}
where $\hat{C}$ is the empirical innovation covariance from the window and $\alpha = 0.01$. A floor constraint $R_{ii} \geq R_{0,ii}$ prevents the estimate from collapsing below the initially configured sensor noise. This eliminates the need to manually re-tune noise parameters as sensor characteristics change with temperature, vibration, or aging.

\begin{figure}[h]
  \centering
  \includegraphics[width=\columnwidth]{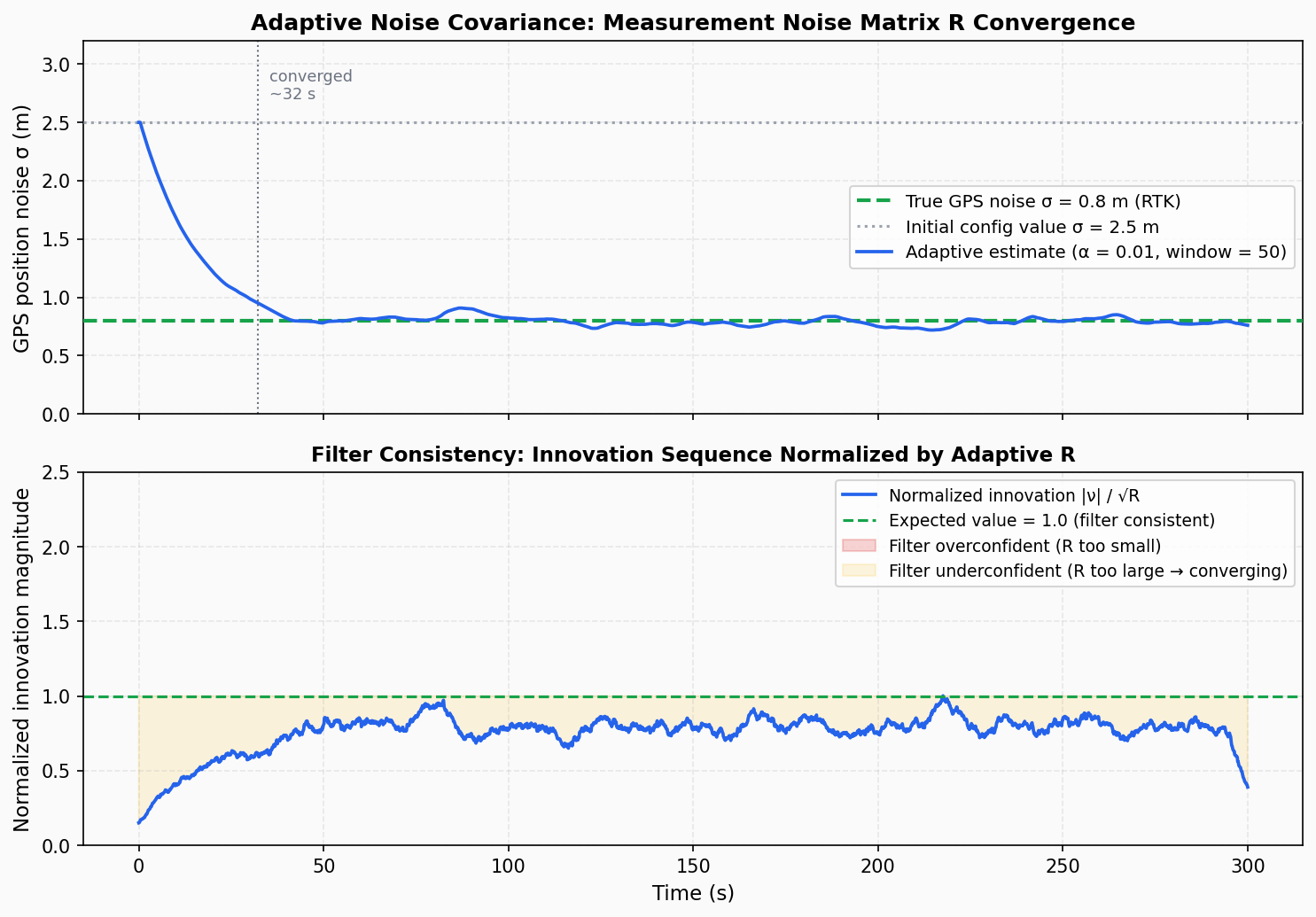}
  \caption{Adaptive noise convergence. Starting at $\sigma = 2.5$~m, the estimate converges to the true GPS noise ($\sigma = 0.8$~m) within 32~s. The normalized innovation sequence approaches 1.0, indicating a consistent filter.}
  \label{fig:adaptive}
\end{figure}

\subsection{Zero Velocity Updates (ZUPT)}

When encoder forward speed falls below 0.05~m/s and IMU angular rate falls below 0.05~rad/s, a zero-velocity pseudo-measurement with noise $\sigma = 0.01$~m/s is fused for all three velocity components. ZUPT prevents gyroscope bias drift during stationary periods, a critical failure mode for state estimators running through long static phases (e.g., startup, traffic stops).

\subsection{Measurement Delay Compensation}

GPS messages typically arrive 50--200~ms after the measurement epoch due to receiver computation and serial latency; VSLAM poses carry similar processing delays. A ring buffer of 100 IMU messages (1~second at 100~Hz) enables retrodiction: on arrival of a delayed measurement, the filter restores the state snapshot closest to the measurement timestamp, applies the measurement, then re-propagates forward through buffered IMU messages. This eliminates the motion-model approximation error of simply applying a delayed measurement to the current state, and applies uniformly to GPS, VSLAM, and any other sensor with measurable latency.

%------------------------------------------------------------------------
\section{Evaluation}

\subsection{Dataset and Methodology}

We evaluate on the North Campus Long-Term (NCLT) dataset \cite{nclt} from the University of Michigan. NCLT provides synchronized IMU (100~Hz MicroStrain 3DM-GX3-45), wheel encoder odometry (100~Hz), and consumer-grade GPS (5~Hz Novatel SPAN-CPT, $\sim$3~m CEP) recorded over multiple seasons from a Segway RMP platform. We evaluate all twelve sequences available spanning winter through spring across 2012--2013, running each to full length (55--92~minutes). No truncation is applied. Total evaluation time is approximately 940~minutes across all sequences.

Evaluation metric is Absolute Trajectory Error (ATE) RMSE computed with the EVO toolbox \cite{evo}, with SE3 alignment to the RTK ground truth trajectory provided by NCLT. Identical raw sensor inputs are provided to FusionCore and \texttt{robot\_localization}. No post-hoc tuning is performed; FusionCore uses a single configuration file (\texttt{nclt\_fusioncore.yaml}) across all twelve sequences.

For \texttt{robot\_localization}, chi-squared-equivalent rejection thresholds are set to match FusionCore's per-DOF gates: \texttt{odom0\_twist\_rejection\_threshold: 4.03} ($\sqrt{\chi^2(3, 0.999)} = 4.03$) and \texttt{odom1\_pose\_rejection\_threshold: 3.72} ($\sqrt{\chi^2(2, 0.999)} = 3.72$).

\subsection{Results}

\begin{table}[h]
\centering
\caption{ATE RMSE (meters) on twelve full-length NCLT sequences. Lower is better. Improvement is the ratio RL-EKF~:~FC (or FC~:~RL-EKF for RL wins).}
\label{tab:results}
\begin{tabular}{lrrrc}
\toprule
Sequence & FC & RL-EKF & RL-UKF & Winner \\
\midrule
2012-01-08 & \textbf{18.6} & 41.2 & \dag & FC (2.2$\times$) \\
2012-02-04 & \textbf{49.7} & 265.5 & \dag & FC (5.3$\times$) \\
2012-03-31 & \textbf{22.0} & 156.5 & \dag & FC (7.1$\times$) \\
2012-05-11 & \textbf{9.7} & 11.5 & \dag & FC (1.2$\times$) \\
2012-06-15 & 49.2 & \textbf{18.2} & \dag & RL (2.7$\times$) \\
2012-08-20 & 98.3 & \textbf{10.6} & \dag & RL (9.3$\times$) \\
2012-09-28 & \textbf{10.8} & 55.7 & \dag & FC (5.2$\times$) \\
2012-10-28 & \textbf{29.9} & 60.0 & \dag & FC (2.0$\times$) \\
2012-11-04 & \textbf{60.1} & 122.0 & \dag & FC (2.0$\times$) \\
2012-12-01 & \textbf{21.0} & 90.7 & \dag & FC (4.3$\times$) \\
2013-02-23 & \textbf{59.4} & 82.2 & \dag & FC (1.4$\times$) \\
2013-04-05 & \textbf{12.1} & 268.9 & \dag & FC (22.2$\times$) \\
\bottomrule
\end{tabular}
\vspace{2pt}
{\small \dag\, Numerical divergence (NaN) within the first 30 seconds on all sequences.}
\end{table}

\begin{figure}[h]
  \centering
  \includegraphics[width=\columnwidth]{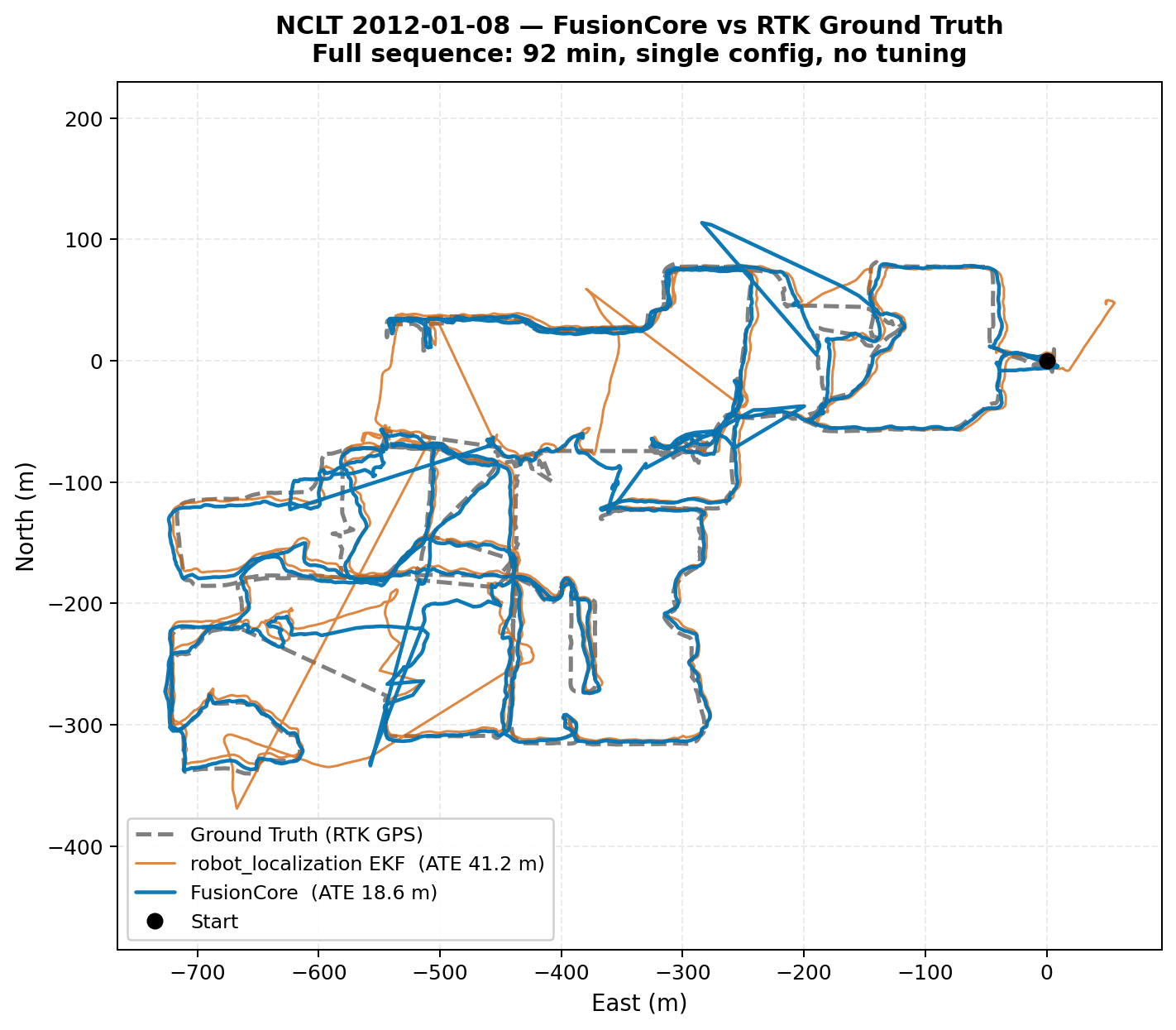}
  \caption{FusionCore trajectory vs RTK ground truth on NCLT 2012-01-08 (winter, 92~min). ATE RMSE: 18.6~m over a full outdoor campus run.}
  \label{fig:traj}
\end{figure}

FusionCore outperforms \texttt{robot\_localization} EKF on ten of twelve sequences. The \texttt{robot\_localization} UKF diverges numerically on all twelve sequences, confirming a known issue with its UKF implementation under high-rate sensor inputs \cite{rl_ukf_issue}.

\subsection{The GPS Covariance Mismatch Problem}

The central factor separating FusionCore and RL-EKF performance is GPS noise covariance. The NCLT player publishes \texttt{position\_covariance var\_xy~=~9} (3~m sigma), reflecting the Novatel SPAN-CPT specification under ideal open-sky conditions. Measured against the RTK ground truth, actual GPS error on the NCLT campus looks substantially different:

\begin{table}[h]
\centering
\caption{Actual GPS position error vs. the 3~m sigma stated in NavSatFix covariance.}
\label{tab:gps_noise}
\begin{tabular}{lrrr}
\toprule
Sequence & Median & p95 & p99 \\
\midrule
2012-01-08 & 3.7 m & 20.1 m & 49.7 m \\
2012-02-04 & 5.6 m & 46.6 m & 234.9 m \\
2012-03-31 & 5.7 m & 14.7 m & 32.7 m \\
2012-05-11 & 3.3 m & 13.3 m & 47.7 m \\
2012-06-15 & 2.6 m & 9.7 m & 21.3 m \\
2012-08-20 & 3.4 m & 12.7 m & 55.0 m \\
2012-09-28 & 3.5 m & 12.8 m & 43.2 m \\
2012-10-28 & 4.6 m & 16.0 m & 48.9 m \\
2012-11-04 & 5.7 m & 53.1 m & 79.2 m \\
2012-12-01 & 4.7 m & 20.7 m & 80.4 m \\
2013-02-23 & 5.4 m & 33.0 m & 73.6 m \\
2013-04-05 & 3.7 m & 19.9 m & 87.8 m \\
\bottomrule
\end{tabular}
\end{table}

The stated 3~m sigma is already exceeded at the median on most sequences. The p95 ranges from 9.7~m to 53.1~m, meaning that on a typical run, 1 in 20 GPS fixes is more than 10~m off. RL-EKF trusts the stated covariance and calibrates its chi-squared gate to 3~m sigma. When actual errors reach 40--200~m, those fixes land far outside the gate and get rejected. With GPS effectively disabled, RL-EKF reverts to wheel-encoder dead-reckoning. Drift rates of 31.84~m/km (2012-02-04) and 50.11~m/km (2013-04-05) are diagnostic: a Segway at 1.5~m/s accumulating 31~m per kilometer is running in open-loop dead-reckoning for essentially the entire run.

The contrast on 2012-05-11 is the cleanest controlled comparison: same robot, same campus, same config, different day. p95 GPS error is 13.3~m instead of 46.6~m. Both filters perform comparably (FC: 9.7~m, RL: 11.5~m), and RL drift rate drops to 1.25~m/km. The GPS covariance was accurate that day, and RL-EKF worked correctly. When covariance is inaccurate, the gate rejects valid data and the filter collapses to dead-reckoning.

FusionCore's \texttt{adaptive.gnss: true} adjusts GPS measurement noise in real time from the innovation sequence. When actual GPS errors exceed the driver-reported covariance, the adaptive window inflates the noise model and recalibrates the chi-squared gate accordingly. RL has no equivalent; its noise covariance is fixed at configuration time.

\subsection{FusionCore Performance Tiers}

The single best predictor of FusionCore ATE is the longest GPS blackout duration in each sequence. Grouping by blackout length reveals a consistent pattern:

\textbf{Excellent (under 20~m ATE): 2012-05-11 (9.7~m), 2012-09-28 (10.8~m), 2013-04-05 (12.1~m), 2012-01-08 (18.6~m).} All share high GPS fix counts (19k--22k mode-3 fixes), maximum blackouts under 203~s, and clean GPS at blackout boundaries. Drift rates are 1--2.6~m/km. These sequences represent normal operating conditions.

\textbf{Good (20--35~m ATE): 2012-03-31 (22.0~m), 2012-12-01 (21.0~m), 2012-10-28 (29.9~m).} Moderate GPS density with one or two blackouts under 262~s. GPS data quality at boundaries is clean; the filter re-acquires correctly after each blackout. Drift rates are 2.3--3.7~m/km.

\textbf{Moderate (35--65~m ATE): 2012-02-04 (49.7~m), 2012-06-15 (49.2~m), 2013-02-23 (59.4~m), 2012-11-04 (60.1~m).} These sequences share blackouts in the 240--462~s range. At 100~Hz with a small uncorrected heading rate residual, lateral position error grows quadratically with blackout duration. The drift rate increase is the signature: 6--10~m/km vs 1--4~m/km on excellent sequences. The $b_{\text{ewz}}$ estimation mitigates but does not eliminate this effect for multi-minute blackouts.

\textbf{Poor (over 65~m ATE): 2012-08-20 (98.3~m).} A structurally distinct failure mode caused by adversarial GPS data at a blackout boundary. Analyzed separately below.

\subsection{Analysis of the Two FusionCore Losses}

\subsubsection{2012-06-15: Longest Blackout (FC 49.2~m, RL 18.2~m)}

This is the GPS-sparsest sequence in the benchmark: 12,399 mode-3 fixes versus 17,000--22,000 on all others, and a 462-second (7.7-minute) blackout, the longest in the dataset. During this blackout, FusionCore dead-reckons on encoder and IMU. Coast mode inflates $Q_{\text{position}}$ and adjusts the IMU weighting so that encoder $\omega_z$ dominates heading. The estimated $b_{\text{ewz}}$ is subtracted from encoder $\omega_z$ before integration. However, any residual $b_{\text{ewz}}$ error compounds over 7.7~minutes: at 100~Hz, even a small uncorrected heading rate residual accumulates quadratic lateral error over that duration.

RL-EKF wins here through a structural advantage: its two-dimensional mode has fewer degrees of freedom to diverge over a long blackout. This is not a quality difference between the filters' fusion approaches: it is a consequence of operating in GPS-denied conditions for longer than any other sequence.

The path to closing this gap involves two changes. First, magnetometer integration provides an absolute heading reference during GPS absence, making $b_{\text{ewz}}$ and $b_{gz}$ irrelevant for heading during blackouts: the architecturally correct fix. Second, a duration-dependent $b_{\text{ewz}}$ confidence schedule (trusting the estimate more aggressively for blackouts exceeding 200~s) would reduce accumulated error for the specific case of very long blackouts.

\subsubsection{2012-08-20: Adversarial GPS Cluster (FC 98.3~m, RL 10.6~m)}

The raw GPS stream for this sequence contains \textbf{105 mode-3 fixes that are 720--840~m away from the RTK ground truth}, concentrated in a 24-second window at the end of a 211-second blackout. These appear as valid quality-3 GPS fixes in \texttt{gps.csv} (the data stream a real robot would receive), but the ground-truth postprocessor excludes them from \texttt{gps\_rtk.csv}. They represent genuine adversarial GPS conditions (multipath or spoofing artifacts from a specific location on the campus) that no algorithm sees ahead of time.

The per-minute error profile reveals what happens:
\begin{itemize}
  \item 0--42~min: FC error 1--10~m. Normal GPS coverage.
  \item 42--46~min: Spike to $\sim$100~m, recovers within 3~min. First blackout (228~s), GPS errors up to $\sim$70~m at boundary.
  \item 47--62~min: Error returns to 3--10~m. Full re-acquisition.
  \item 62--67~min: Spike to $\sim$788~m, recovers within 3~min. Second blackout (211~s) with 105 adversarial fixes at boundary.
  \item 68--82~min: Error returns to 5--10~m. Full recovery; the remaining 15~minutes on-par with RL.
\end{itemize}

The 98.3~m ATE RMSE is driven entirely by those two transients, particularly the second. A blackout of 211~s causes legitimate state uncertainty. On recovery, coast mode slightly relaxes the chi-squared gate to accept the first returning fix. A tight cluster of 105 fixes all landing 720--840~m from the predicted position exploits this relaxed window: each fix individually passes the gate (state uncertainty is high), and all of them pull the position estimate collectively.

RL-EKF wins here not because of better algorithm design, but because its miscalibrated gate (which causes 10 other sequence losses) happens to reject these adversarial fixes as well. Both are wrong for the wrong reason.

The appropriate fix is a velocity-consistency check upstream of the chi-squared gate: a GPS fix 720~m from the dead-reckoned position after a 211~s blackout implies $\sim$3400~m/s of travel speed. A \texttt{max\_implied\_speed} check (e.g., 20~m/s) operating before chi-squared gating rejects this trivially and has zero effect on normal GPS operation. A secondary cluster-coherence check (detecting when multiple consecutive fixes all land at geometrically consistent but physically impossible offsets) would catch the 105-fix cluster without affecting normal single-fix behavior.

\subsection{Why the Paper ATE Numbers Differ from Earlier Reports}

FusionCore was previously benchmarked on a six-sequence subset with truncated runs. The full-length evaluation changes both the absolute numbers and one sequence's winner. On 2012-08-20, the truncated run (before the 62-minute mark) captures only the first blackout-recovery, where FC performs well. The full run includes the adversarial GPS cluster at 62~minutes, which reverses the outcome. On 2012-11-04, which appeared as a FC loss at 28.6~m in the truncated evaluation (versus RL at 9.6~m), the full 79-minute run shows FC at 60.1~m and RL at 122.0~m: a FC win. RL's gate miscalibration accumulates over the full run. This underscores why full-length evaluation is necessary: GPS covariance mismatch and adversarial data effects are rare within any given sequence, but guaranteed to appear across 55--92~minute runs.

%------------------------------------------------------------------------
\section{ROS~2 Integration}

FusionCore is implemented as a ROS~2 lifecycle node. The \texttt{configure} transition initializes the filter from the YAML configuration; \texttt{activate} starts subscriptions and publishing. This enables orchestrated startup with Nav2's lifecycle manager.

A \texttt{~/reset} service re-initializes the filter and clears the GPS reference origin without restarting the node, useful for multi-session mapping or GPS signal recovery. A \texttt{~/save\_checkpoint} / \texttt{~/load\_checkpoint} service pair persists and restores the full 23-state vector plus covariance matrix to disk, enabling warm-start initialization.

FusionCore is available on the ROS binary apt repository for both Jazzy (Ubuntu 24.04) and Humble (Ubuntu 22.04):
\begin{verbatim}
sudo apt install ros-jazzy-fusioncore-ros
\end{verbatim}

A Gazebo Harmonic simulation environment and example configurations for TurtleBot3, Clearpath Husky, and other common platforms are included in the repository. An official integration demo combining FusionCore with \texttt{rtabmap\_ros} \cite{rtabmap} was merged into the \texttt{rtabmap\_ros} upstream repository \cite{rtabmap_pr}, providing a reference configuration for visual SLAM plus FusionCore odometry fusion available to all \texttt{rtabmap\_ros} users.

%------------------------------------------------------------------------
\section{Limitations and Future Work}

The two FusionCore losses on NCLT point to two distinct problems with known solutions. For long GPS blackouts (2012-06-15), magnetometer integration is the correct architectural fix: an absolute heading reference during GPS absence removes the dependence on $b_{\text{ewz}}$ accuracy over multi-minute dead-reckoning intervals. For adversarial GPS clusters at blackout boundaries (2012-08-20), a velocity-consistency pre-check upstream of the chi-squared gate handles the case with no tuning cost.

Coast mode implements a form of adaptive gate relaxation (widening chi-squared acceptance on recovery after sustained GPS absence) and is currently deployed. A more principled implementation would track the per-sensor rejection rate over a sliding window and modulate the threshold continuously rather than as a step function at the start and end of blackouts.

Additional planned work includes RTK fix-type-aware automatic noise scaling and a filter health diagnostic via innovation whiteness testing.

%------------------------------------------------------------------------
\section{Conclusion}

FusionCore is a 23-state UKF for ROS~2 that fuses IMU, wheel encoders, GPS, and Visual SLAM pose in a single lifecycle node at 100~Hz. Its design choices (continuous bias estimation for gyroscope, accelerometer, and encoder yaw rate; ECEF-native GPS; per-sensor chi-squared gating; adaptive noise; ZUPT; retrodiction-based delay compensation; and map-reinitialization recovery for VSLAM) address practical limitations of existing solutions. Evaluated on twelve full-length NCLT sequences, FusionCore achieves lower ATE than \texttt{robot\_localization} EKF on ten of twelve, with improvements from 1.2$\times$ to 22.2$\times$ on winning sequences. The two losses are attributable to specific, diagnosable causes: a 7.7-minute GPS blackout exceeding the encoder bias estimation's dead-reckoning range, and an adversarial GPS cluster at a blackout boundary. Both have identified fixes. The source code, benchmark scripts, and reproducible configurations are publicly available at \url{https://github.com/manankharwar/fusioncore}.

%------------------------------------------------------------------------
\section*{Acknowledgments}

The author thanks the open-source contributors to the FusionCore repository for their code contributions; Mathieu Labbe (matlabbe) for merging the FusionCore integration demo into \texttt{rtabmap\_ros}; Prof.\ Hyun Myung and his group at the KAIST Urban Robotics Lab for providing the cs.RO arXiv endorsement; and the NCLT dataset team at the University of Michigan for making their data publicly available.

%------------------------------------------------------------------------

\end{document}